\documentclass[runningheads]{llncs}
\usepackage{graphicx}
\usepackage{amsmath,amssymb} 
\usepackage{color}
\begin{document}
\title{Multi-modal Segment Assemblage Network\\for Ad Video Editing with\\Importance-Coherence Reward}
\titlerunning{M-SAN with Imp-Coh Reward for Ad Video Editing}
%
%
\author{Yunlong Tang\inst{1,2}\orcidID{0000-0003-2796-1787} \and
Siting Xu\inst{1}\orcidID{0000-0001-9934-7919} \and\\
Teng Wang\inst{1,2} \and
Qin Lin\inst{2} \and
Qinglin Lu\inst{2} \and
Feng Zheng\inst{1}\orcidID{0000-0002-1701-9141}\thanks{Corresponding author}
}
\authorrunning{Y. Tang et al.}
%
\institute{Southern University of Science and Technology, China \and
Tencent Inc., China\\
\email{\{tangyl2019, xust2019, wangt2020\}@mail.sustech.edu.cn}\\
\email{\{angelqlin, qinglinlu\}@tencent.com}\qquad
\email{f.zheng@ieee.org}
}
\maketitle              
\begin{abstract}
Advertisement video editing aims to automatically edit advertising videos into shorter videos while retaining coherent content and crucial information conveyed by advertisers. It mainly contains two stages: video segmentation and segment assemblage. The existing method performs well at video segmentation stages but suffers from the problems of dependencies on extra cumbersome models and poor performance at the segment assemblage stage. To address these problems, we propose M-SAN (Multi-modal Segment Assemblage Network) which can perform efficient and coherent segment assemblage task end-to-end. It utilizes multi-modal representation extracted from the segments and follows the Encoder-Decoder Ptr-Net framework with the Attention mechanism. Importance-coherence reward is designed for training M-SAN. We experiment on the Ads-1k dataset with 1000+ videos under rich ad scenarios collected from advertisers. To evaluate the methods, we propose a unified metric, Imp-Coh@Time, which comprehensively assesses the importance, coherence, and duration of the outputs at the same time. Experimental results show that our method achieves better performance than random selection and the previous method on the metric. Ablation experiments further verify that multi-modal representation and importance-coherence reward significantly improve the performance. Ads-1k dataset is available at: https://github.com/yunlong10/Ads-1k

\keywords{Ad Video Editing  \and Segment Assemblage \and Advertisement Dataset \and Multi-modal \and Video Segmentation \and Video Summarization.}
\end{abstract}
\section{Introduction}
With the boom of the online video industry, video advertising has become popular with advertisers. However, different online video platforms have different requirements for the content and duration of ad videos. It is time-consuming and laborious for advertisers to edit their ad videos into a variety of duration tailored to the diverse requirements, during which they have to consider which part is important and whether the result is coherent. Therefore, it is of great importance to automatically edit the ad videos to meet the requirements of duration, and the edited videos should be coherent and retain informative content.

\begin{figure}
\centering
\includegraphics[width=100mm]{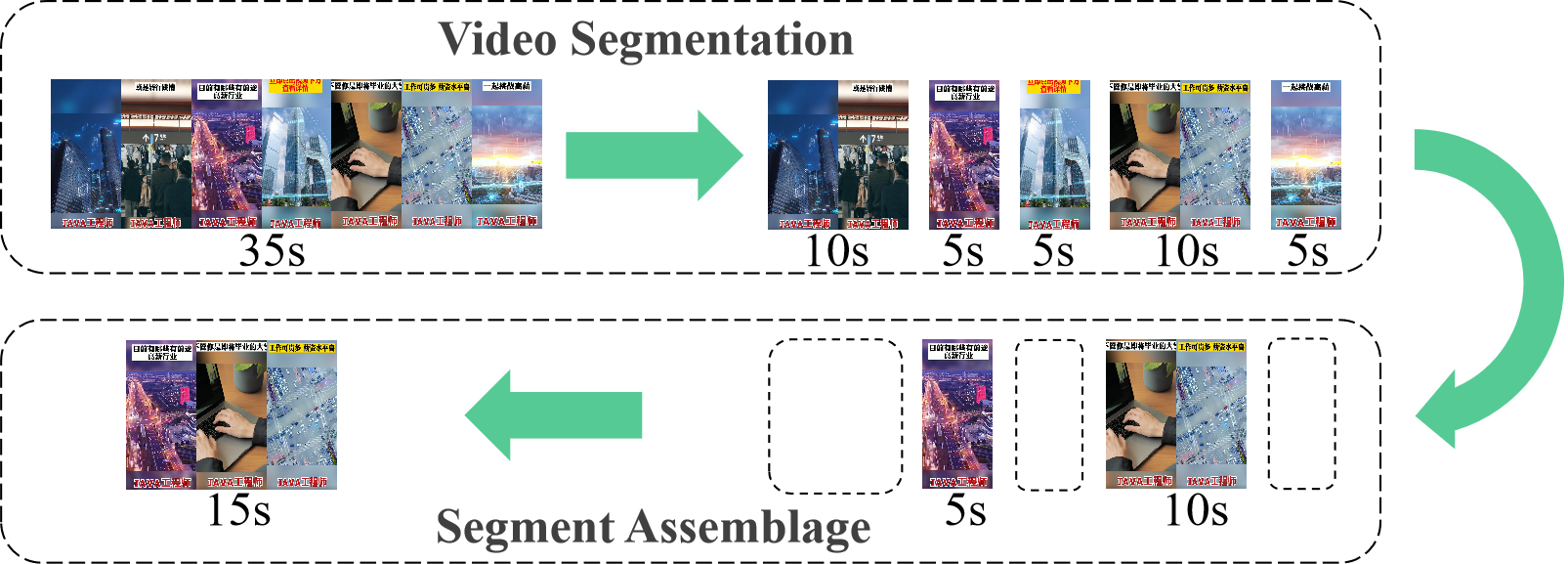} 
\caption{The two stages of ad video editing: video segmentation and segment assemblage.}
\label{fig:shiyi}
\end{figure}

Ad video editing is a task aiming to edit an ad video into its shorter version to meet the duration requirements, ensuring coherence and avoiding losing important ad-related information. Video segmentation and segment assemblage are the two main stages in ad video editing task \cite{MVSM}, as Fig. \ref{fig:shiyi} shows. An ad video will be cut into several segments with a small duration during the video segmentation stage. At the segment assemblage stage, the output will be produced by selecting and assembling a subset of the input segments of the source ad video. The key to video segmentation is to preserve the local semantic integrity of each video segment. For instance, a complete sentence of a speech or caption in the source video should not be split into two video segments. Existing method \cite{MVSM} has achieved this by aligning shots, subtitles, and sentences to form the segments. However, at the segment assemblage stage, the only pioneer work \cite{MVSM} suffers the following problems: (1) To calculate the individual importance of each segment and the coherence between segments, extra models are required to perform video classification \cite{nextvlad,vggish} and text coherence prediction \cite{bert}, which is inefficient during inference. (2) Without globally modeling the context of videos, graph-based search adopted by \cite{MVSM} produces results with incoherent segments or irrelevant details.

To tackle these problems, we propose an end-to-end  Multi-modal Segment Assemblage Network (M-SAN) for accurate and efficient segment assemblage. It is free of extra cumbersome models during inference and strikes a better balance between importance and coherence. Specifically, we obtain segments at the video segmentation stage by boundary detection and alignment. Different from daily life videos, ad videos usually have sufficient multi-modal content like speech and caption, which contain abundant video semantics \cite{guo}. Therefore, pretrained unimodal models are applied to respectively extract the representation of shots, audios, and sentences, which are concatenated together yielding a multi-modal representation of segments. During the assemblage stage, we adopt a pointer network with RNN-based decoder to improve the temporal dependency between selected segments. Importance-coherence reward is designed for training M-SAN with Policy Gradient \cite{pg}. Importance reward measures the amount of important ad-related information contained in the output. Coherence reward measures the text coherence between every two adjacent selected segments, which is computed as the mean of PPL (perplexity) \cite{gpt2} of sentences generated by concatenating the two texts extracted from adjacent selected segments.

To evaluate our methods, we propose the new metric, Imp-Coh@Time, which takes the importance, coherence, and duration of the outputs into consideration at the same time instead of evaluating importance or coherence respectively. We experiment on the Ads-1k dataset with 1000+ ad videos collected from advertisers. Experimental results show that our method achieves better performance than random selection and the previous method \cite{MVSM} on the metric. Ablation experiments further verify that multi-modal representation and importance-coherence reward significantly improve the performance.

Our work mainly focuses on segment assemblage in ad video editing, and its main contributions can be summarized as follows:

\begin{itemize}
    \item[$\bullet$] We propose M-SAN to perform segment assemblage efficiently and improve the result of ad video editing, without relying on an extra model when inference.
    \item[$\bullet$] We propose importance-coherence reward and train M-SAN with policy gradient to achieve a better trade-off between importance and coherence.
    \item[$\bullet$] We collect the dataset Ads-1k with 1000+ ad videos and propose Imp-Coh@Time metrics to evaluate the performance of ad video editing methods. Our M-SAN achieves state of the art on the metrics.
\end{itemize}

\section{Related Work}

\subsection{Video Editing}
There are three main categories of automated video editing \cite{MVSM}. They're video summarization, video highlight detection, and task-specific automated editing.\\
\textbf{Video Summarization.} The most relevant task to ad video editing is video summarization. It is a process that extracts meaningful shots or frames from video by analyzing structures of the video and time-space redundancy in an automatic or semi-automatic way. To perform video summarization, a load of work focuses on supervised learning based on frames \cite{frame1,frame2,frame3,frame4}, shots \cite{shot1,shot2,shot3}, and clips \cite{clip}. Other than these works, DSN \cite{DSN} is the first proposed unsupervised video summarization model training with diversity-representativeness reward by policy gradient. Without utilizing annotation, the method reached fully unsupervised. Our rewards design mainly refers to \cite{DSN}.\\
\textbf{Video Highlight Detection.} Learning how to extract an important segment from videos is the main motivation we focus on video highlight detection.  \cite{highlight3} proposed frameworks that exploit users' previously created history. Edited videos created by users are utilized in  \cite{highlight1} to achieve highlight detection in an unsupervised way since human-edited videos tend to show more interesting or important scenes.  \cite{highlight4} presents an idea that shorter videos tend to be more likely to be selected as highlights. Combining the above ideas, we exploit the ability of MSVM \cite{MVSM} which extracts potential selling points in segments that tend to be of higher importance when performing assemblage.\\
\textbf{Task-specific Automated Editing.} Videos can be presented in various forms such as movies, advertisements, sports videos, etc. To extract a short video from a long video also exists in specific scenarios of movies  \cite{movies,movies2,movies3}.  \cite{movies} pointed out that a movie has the trait that its computational cost is high. Also, sports videos have been explored to extract highlights \cite{sports,sports2} for sports having the characteristic that they are high excitement. Advertisements are rich in content and they vary in duration among different platforms \cite{MVSM}.

\subsection{Text Coherence Prediction and Evaluation}
Text information is extracted in the stage of video segmentation. When assembling segments, texts that are concatenated act as a reference to coherence.  \cite{coherence1} proposed narrative incoherence detection, denoted semantic discrepancy exists causes incoherence. In  \cite{MVSM}, next sentence prediction (NSP) \cite{bert} is exploited to assess coherence. Although our work is inspired by their thoughts, we use perplexity as the coherence reward and metric to evaluate sentence coherence.

\subsection{Neural Combinatorial Optimization}
Combinatorial optimization problem is a problem that gets extremum in discrete states. Common problems like Knapsack Problem (KP), Travelling Salesman Problem (TSP), and Vehicular Routing Problem (VRP) belong to combinatorial optimization problem. Pointer Network (Ptr-Net) is proposed in  \cite{ptrnet} and performs better than heuristic algorithm in solving TSP. Later, Ptr-Net has been exploited with reinforcement learning \cite{ptrreinforce1,ptrreinforce2,ptrreinforce3,ptrreinforce4,exactk}. In  \cite{ptr}, Ptr-Net is used to solve the length-inconsistency problem in video summarization. In our network, Ptr-Net is used to model the video context and select the tokens from the input sequence as output.


\section{Method}
\subsection{Problem Formulation}
 Given a set of $N$ segments of ad video $S=\{s_i\}_{1\leq i\leq M}$, our goal is to select a subset $A=\{a_i\}_{1\leq i\leq N}\subseteq S$ which can be combined into the output video so that it can take the most chance to retain the important information and be coherent as well as meeting the requirements of duration. We denote the ad importance of the segment $a_i$ as $imp(a_{i})$, the coherence as $coh(a_{i})$ and the duration as $dur(a_{i})$. Overall the task of segment assemblage can be regarded as a constrained combinatorial optimization problem and is defined formally as follows:
 \begin{equation}
     \begin{split}
         \max_{A\subseteq S}~\sum_{a_{i}\in A}imp(a_{i})&+\sum_{a_{i}\prec a_{j}}coh(a_{i},a_{j})~,\\s.t.~T_{min}\leq\tau(A)\leq T_{max},&~and~
\forall a_{i},a_{j}\in A, a_{i}\neq a_{j}~,
     \end{split}
 \end{equation}
where
\begin{itemize}
    \item[$\bullet$] $T_{min}$ and $T_{max}$ are the lower and upper bound of requirement duration,
    \item [$\bullet$] $\tau(A)=\sum_{a_i\in A}dur(a_{i})$,
    \item [$\bullet$] $a_i\prec a_j$ is defined as $(a_i,a_j\in A)\land(i<j)\land(\forall a_k\in A,~k<i\lor k>j)$. 
\end{itemize}
This is an NP-hard problem that can not be solved in polynomial time. Instead of utilizing graph modeling and optimization \cite{MVSM} to search for an optimal solution, we adopt a neural network with pointer \cite{ptrnet} that follows the framework of neural combinatorial optimization to optimize the objective directly.
\subsection{Architecture}
\begin{figure}
\centering
\includegraphics[width=120mm]{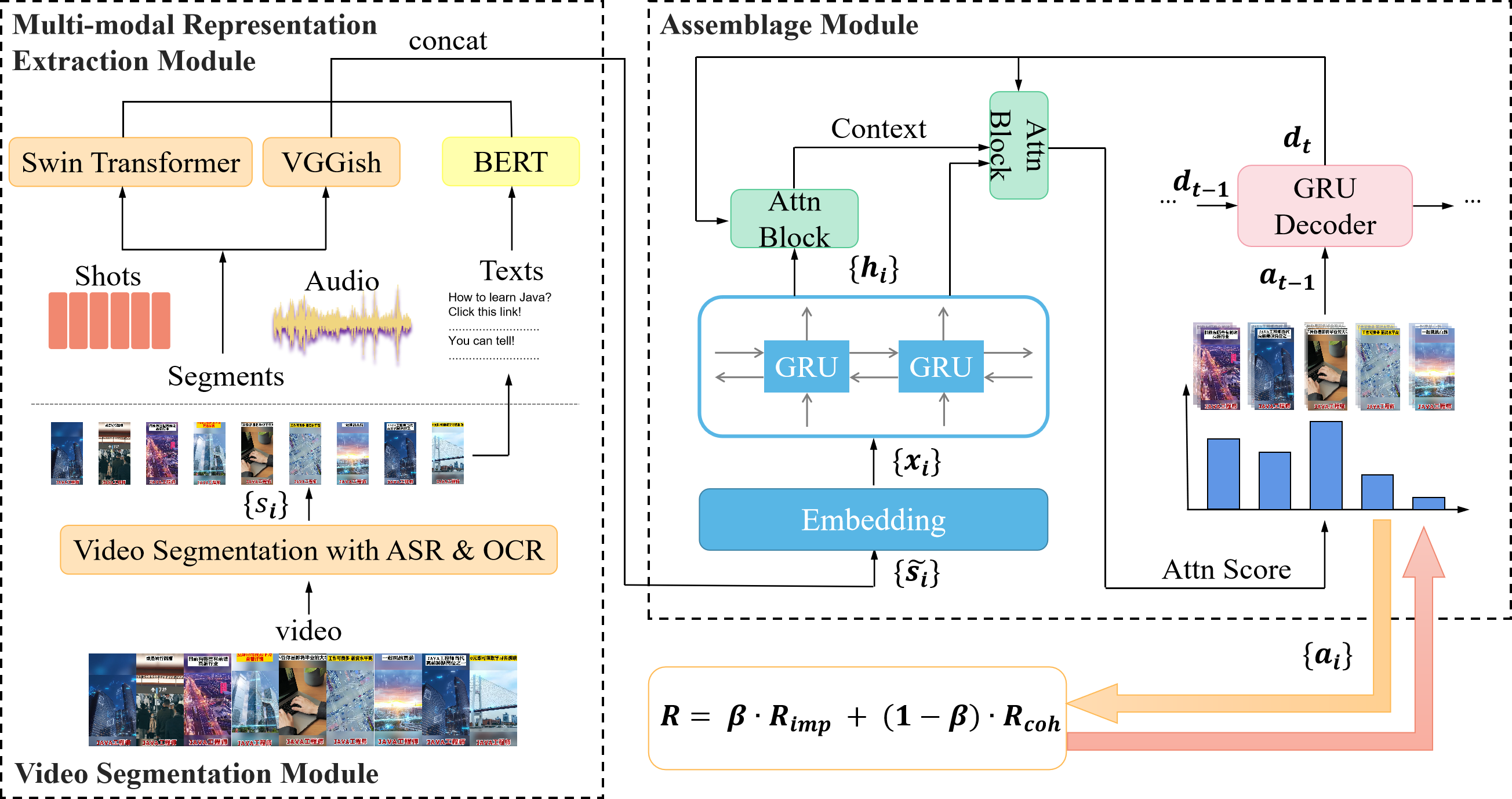} 
\caption{The Architecture of M-SAN}
\label{fig:model}
\end{figure}

The architecture of M-SAN is shown as Fig. \ref{fig:model}. It incorporates a multi-modal video segmentation module \cite{MVSM} (MVSM), multi-modal representation extraction module (MREM) and assemblage module (AM). To preserve the local semantic integrity of each segment, we adopt MVSM to obtain video segments with reasonable boundaries. With ASR and OCR, MVSM also captures the texts from each segment. Given the segments and the corresponding texts, MREM extracts the segment-level representations of shots, audios, and texts, which are further jointed into the multi-modal representations. AM utilizes these representations to model the context of video and make decisions by a pointer network \cite{ptrnet}.
\subsubsection{Video Segmentation Module.}
At the video segmentation stage, we first apply MVSM \cite{MVSM} to obtain the segments of each input video. MVSM splits a video into the video track and audio track and extracts shots $\{v_i\}$, audios $\{\alpha_i\}$, ASR and OCR results $\Omega_i=\{\omega_i\}$ (a sentence with words $\omega_i$) from video to generate the boundaries of the content in each modality. The boundaries of audio space and textual space are first merged to form the joint space, followed by merging the boundaries of visual space and joint space to yield the final segments set $\{s_i\}$. The segment $s_i=(\{v_{p_0},..., v_{q_0}\},\{\alpha_{p_1},...,\alpha_{q_1}\},\Omega_i)$ preserves the integrity of local atomic semantic, where $p_{(\cdot)}<q_{(\cdot)}$.
\subsubsection{Multi-modal Representation Extraction Module.}
MREM integrates three kinds of representation extractors: pre-trained Swin-Transformer \cite{swin}, Vggish \cite{vggish} and BERT \cite{bert} models. The Swin-Transformer extracts the visual representations $\{\tilde v_{p_0},...,\tilde v_{q_0}\}$ from shots $\{v_{p_0},..., v_{q_0}\}$ in each segment $\{s_i\}$, and a segment-level visual representation is computed as the mean of single shot-level representaton. Vggish and BERT model extract audio representations $\{\tilde\alpha_{p_1},..., \tilde\alpha_{q_1}\}$ and text representations $\tilde\Omega_i$, respectively. Similarly, a segment-level audio representation $\bar\alpha_i$ is given by the mean of $\{\tilde\alpha_{p_1},..., \tilde\alpha_{q_1}\}$. Since each segment contains at most one sentence, it is $\tilde\Omega_i$ that is the segment-level text representation. These three modalities will be jointed by concatenating directly to yield the final multi-modal representation of segment $\tilde s_i=[\bar v_i,\bar\alpha_i, \tilde\Omega_i]^T$.

\subsubsection{Assemblage Module.}
\label{AM}
The output of an Encoder-Decoder Ptr-Net \cite{ptrnet} is produced by iteratively copying an input item that is chosen by the pointer \cite{DeepQAMVS}, which is quite suitable for segment assemblage task. Therefore, our assemblage module (AM) follows this framework.

The encoder integrates a linear embedding layer and a bi-directional GRU. To enhance interactions between modalities, the linear embedding layer will perform preliminary a fusion of the three modalities and produce embedding token $x_i$ corresponding to segment $s_i$. To enhance the interaction between segments and model the context of the whole video, a Bi-GRU is adopted to further embed the tokens:
\begin{equation}
H = GRU_{e}(X)~,
\end{equation}
where $X=[x_1,...,x_M]$, $H=[h_1, ..., h_M]$, and the hidden state $h_i$ is the context embedding for segment $s_i$.

Given the output of encoder $H$ and $X$, the GRU \cite{gru} decoder with Attention mechanism \cite{att-m} predicts the probability distribution of segment to be selected from $S$ at every time-step $t$ to get the result $A$:
\begin{equation}
    p_\theta(A|S)=\prod_{t=1}^{N}p_\theta(a_t|a_{1:t},S)=\prod_{t=1}^{N}p_\theta(a_t|a_{1:t-1},H,X)=\prod_{t=1}^{N}p_\theta(a_t|a_{t-1},d_t)~,
\end{equation}
where $\theta$ is the learnable parameter, and $d_t$ is the hidden state computed by the decoder at time-step $t$. With $d_t$ as the query vector, the decode will glimpse \cite{glimpse} the whole output of encoder $H$ to compute the bilinear attention. Instead of utilizing the additive attention adopted in  \cite{ptr,exactk}, we compute bilinear attention $\mu_t$ with less computational cost:
\begin{equation}
\mu_t = Softmax(H^TW_{att_1}d_t)~.
\end{equation}
Until now, the attention $\mu_t$ is used as the probability distribution to guide the selection in most of Ptr-Net framework \cite{ptr,exactk}. To dynamically integrate information over the whole video \cite{glimpse2}, we further calculate the context vector $c_t$ of encoder output and update query vector $d_t$ to $\tilde d_t$ by concatenating it with $c_t$ (Eq. \ref{eq:d}). Then we compute attention a second time to obtain the probability distribution of segment selection at current time-step $t$ (Eq. \ref{eq:first_att}).
\begin{equation}
\label{eq:d}
c_t=H\mu_t=\sum_{m=1}^{M}\mu_t^{(m)}h_m,~~~\tilde d_t = 
\begin{bmatrix}
d_t\\
c_t
\end{bmatrix}~,
\end{equation}
\begin{equation}
\label{eq:first_att}
\tilde\mu_t = Softmax(MH^TW_{att_2}\tilde d_t)~,
\end{equation}
\begin{equation}
p_\theta(a_t|a_{t-1},d_t)=\tilde\mu_t~,
\end{equation}
where $M$ can mask the position $i$ corresponding to selected segment $a_i$ to $-\infty$. The segment at this position will be not selected at later time-steps, since $Softmax(\cdot)$ modifies the corresponding probability to 0. Finally, the segment selected at time-step $t$ will be sampled from the distribution:
\begin{equation}
a_t\sim p_\theta(a_t|a_{t-1},d_t)~.
\end{equation}
If the sum of duration of selected segments $\tau(A)$ exceeds the tolerable duration limit $[T_{min},T_{max}]$, the current and following segments selected will be replaced by [EOS] token \cite{len_ctrl}.
\subsection{Reward Design}
\subsubsection{Importance Reward.} To extract important parts from original ad videos, rewards related to the importance of selected segments should be fed back to the network during training. We design importance reward $R_{imp}$:
\begin{equation}
\label{eq:def_R_imp}
    R_{imp}=\frac{1}{|A|}\sum_{a_i\in A}imp(a_i)~,
\end{equation}
\begin{equation}
\label{eq:def_imp}
    imp(a_i)=\frac{1}{|L_{a_i}|}\sum_{\ell\in L_{a_i}}w_{l}\cdot \ell~,
\end{equation}
where A is the set of selected segments, $L_{a_i}$ stands for the total number of labels of the selected segment, $\ell\in\{1,2,3,4\}$ is narrative techniques label hierarchy, and $w_{l}$ is the weight of one label. According to Eq. \ref{eq:def_R_imp}, the importance of a single segment is the weighted average of its annotated labels' weight. The labels listed in the supplementary are divided into four groups with four levels ranging from 1 to 4 according to their ad-relevance. We compute the importance reward of one output as the mean of the importance of single segments.

\subsubsection{Coherence Reward.}
Importance reward mainly focuses on local visual dymantics within the single segment, neglecting temporal relationships between adjacent segments. We introduce a linguistic coherence reward to improve the fluency of caption descriptions of two segments.

\begin{figure}
\centering
\includegraphics[width=120mm]{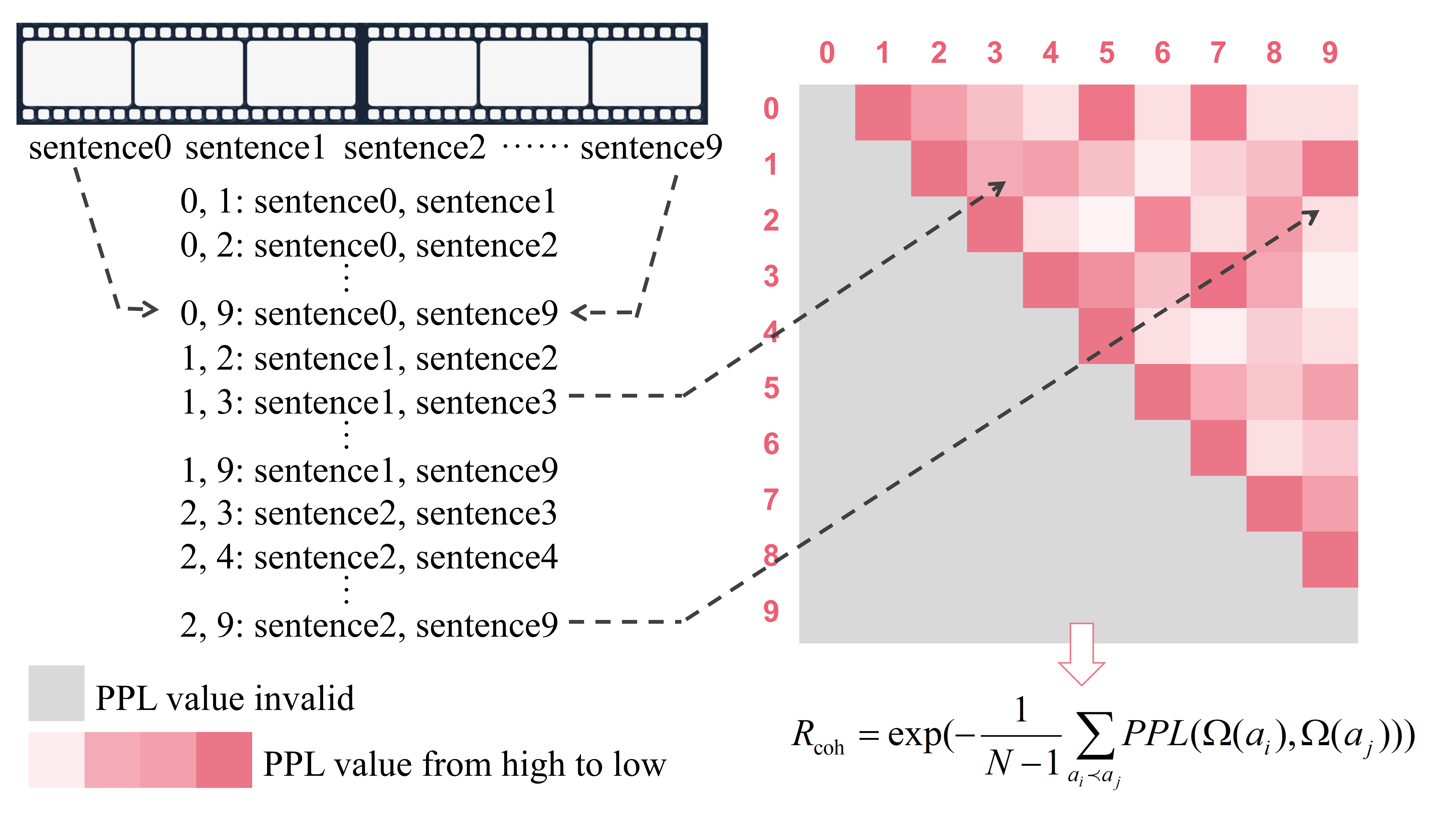}
\caption{PPL map and coherence reward}
\label{fig:ppl}
\end{figure}

Specifically, the extracted texts from adjacent segments are combined in pairs while retaining the original order, that is, preserving the same order of the original precedence. Then we compute the perplexity \cite{gpt2} (PPL) for each combined sentence by GPT-2 pre-trained on 5.4M advertising texts:
\begin{equation}
    \begin{aligned}
        PPL(\Omega_{1},\Omega_{2})&=p(\omega^{(1)}_1,\omega^{(1)}_2,...,\omega^{(1)}_m,\omega^{(2)}_1,\omega^{(2)}_2,...,\omega^{(2)}_n)^{-1/(m+n)}\\
        &=\sqrt[m+n]{\prod_{i=1}^{m+n} \frac{1}{p(\omega_i|\omega_1,\omega_2,...,\omega_{i-1})}}~,
    \end{aligned}
\end{equation}
where $\Omega_i=(\omega_1^{(i)},...,\omega_m^{(i)})$ is one sentence with $m$ words. PPL reflects the \textit{incoherence} of a sentence. We also maintain a PPL map to store the PPL of sentences in pairs as Fig. \ref{fig:ppl} shown.
The grey parts represent invalid PPL values, which means sentence pairs that violate the original orders have no valid PPL value. The rest means valid PPL values vary from high to low as the color of red deepens.
Since the smaller the PPL, the better the text coherence, the coherence reward is computed with a transfer function:
\begin{equation}
\label{eq:def_R_coh}
    R_{coh} = exp(-\frac{1}{N-1}\sum_{a_i\prec a_j} PPL(\Omega(a_i),\Omega(a_j)))~,
\end{equation}
where $N=|A|$ is the total number of selected segments, $\Omega(a_i)$ is the sentence recognized from segment $a_i$. The $exp(\cdot)$ ensures the $R_{coh}$ at the same order of magnitude with $R_{imp}$.

\subsubsection{Importance-Coherence Reward.}
To balance the importance and coherence of the selection, we make $R_{imp}$ and $R_{coh}$ complement each other and jointly guide the learning of M-SAN:
\begin{equation}
\label{reward}
    R = \beta\cdot R_{imp} + (1-\beta)\cdot R_{coh}~,
\end{equation}
where the coefficient of reward $\beta$ is a hyperparameter.
\subsection{Training}
Policy gradient (a.k.a. REINFORCE algorithm \cite{pg}) is adopted during training:



\begin{equation}
\label{REIN_b}
    \begin{aligned}
        \nabla_{\theta}J(\theta) \approx \frac{1}{K}\sum_{k=1}^K\sum_{t=1}^N (R(A^{(k)})-b^{(k)})\nabla_{\theta}\log p_{\theta}(a_t^{(k)}|a_{t-1}^{(k)},d_t^{(k)})~,
    \end{aligned}
\end{equation}
where $K$ denotes the number of episodes, $a_t$ stands for the actions (which segment to choose) and $d_t$ is the hidden state estimated by the decoder. $R(A)$ is the reward calculated by Eq.\ref{reward}. A baseline value $b$ so that the variance can be reduced. For the optimization, neural network parameter $\theta$ is updated as:
\begin{equation}
\label{optimize}
    \begin{aligned}
        \theta = \theta - \eta{\nabla_{\theta}}(-J(\theta))~,
    \end{aligned}
\end{equation}
where $\eta$ is the learning rate.
\section{Experiments}
\subsection{Dataset}
\label{Section:Dataset}
In \cite{MVSM}, there are only 50 ad videos used in experiments. To obtain better training and evaluation, we collect 1000+ ad videos from the advertisers to form the Ads-1k dataset. There are 942 ad videos for training and 99 for evaluation in total. However, the annotation methods of the training set and test set are somehow different. Instead of preparing the ground-truth for each data, we annotate each video with multi-labels shown in the supplementary.

\begin{table}
\begin{center}
\caption{Dataset statistics. $N_{seg}$ and $N_{label}$ are respectively the average number of segments and labels of each video. $D_{seg}$ and $D_{video}$ are the average duration of a segment and a video, respectively.}
\label{tab:dataset1}
\begin{tabular}{c  c  c  c  c}
\hline
 Dataset & $N_{seg}$ & $D_{seg}(s)$ & $D_{video}(s)$ & $N_{label}$ \\ 
\hline
 Training Set & ~~~13.90~~~ & ~~~2.77~~~ & ~~~34.60~~~ & ~~~30.18~~~ \\  
 Test Set & ~~~18.81~~~ & 1.88 & 34.21 & 35.77   \\
 Overall & 14.37 & 2.68 & 35.17 & 30.71\\
\hline
\end{tabular}
\end{center}
\end{table}

We counted the average segment number for videos, the average length of segments in seconds, the average duration of each video in seconds, and the average label number for each video for the training set, test set, and the whole dataset respectively. The results are shown as Tbl. \ref{tab:dataset1}. Besides, the number of annotated segment pairs and the proportion are counted. The number of \textit{coherent}, \textit{incoherent}, and \textit{uncertain} pairs are 6988, 9551, and 2971, occupying 36\%, 49\%, and 15\%, respectively.

\subsection{Metric}
\label{Section:Metric}
\subsubsection{Imp@T.}
 We define the score of ad importance given the target duration $T$ as follows:
 \begin{equation}
 \label{eq:def_imp@T}
    Imp@T=\frac{1}{|A|}\sum_{a_i\in A}imp(a_i)\cdot\mathbb{I}[c_1\cdot T\leq\tau(A)\leq c_2\cdot T]~,
\end{equation}
where A is the set of selected segments, $imp(a_i)$ is defined by Eq. \ref{eq:def_imp}. $\mathbb{I}(\cdot)$ is the indicator function. $c_1$ and $c_2$ are two constant that produced a interval based on given target duration $T$. We set $c_1=0.8$ and $c_2=1.2$, since a post-processing of $0.8\times$ slow down $1.2\times$ or fast forward can resize the result close to the target $T$ without distortion in practice. Take $T=10$ as example, the interval will be $[8,12]$, which means if the duration of result $\tau(A)=\sum_{a_i\in A}dur(a_i)\in [8,12]$ then this result is valid and gain the score.
\subsubsection{Coh@T.}
The coherence score given the target duration $T$ is defined as follows:
\begin{equation}
Coh@T=\frac{1}{|A|-1}{\sum_{a_i\prec a_j}coh{(a_{i},a_{j})}}\cdot\mathbb{I}[c_1\cdot T\leq\tau(A)\leq c_2\cdot T]~,
\end{equation}where the $coh{(a_{i},a_{j})}$ is the coherence small score between the text of segment $i$ and the text of segment $j$. With the annotation for coherence on test set, we can score the results produced by our models. When scoring, for each combination of consecutive two segments $i$ and $j$ in output, if it is in the \textit{coherent} set, the $coh{(a_{i},a_{j})}$ will be 1. If it is in the \textit{incoherent} set, the $coh{(a_{i},a_{j})}$ will be 0. Otherwise, it is in the \textit{uncertain} set, the $coh{(a_{i},a_{j})}$ will be 0.5.
\subsubsection{Imp-Coh@T.}
The overall score is defined as follows:
\begin{equation}
    ImpCoh@T=\frac{Imp@T}{|A|-1}\cdot{\sum_{a_i\prec a_j}coh{(a_{i},a_{j})}}~,
\end{equation}
where Imp@T is defined in Eq. \ref{eq:def_imp@T}. The score reflects the ability of trade-off among importance, coherence and total duration.

\subsection{Implementation Details}
\subsubsection{Baselines.}Besides SAM (Segments Assemblage Module) proposed in  \cite{MVSM}, we also adopt two random methods to perform the segment assemblage task,  given the segments produced by MVSM \cite{MVSM}.
\begin{itemize}
    \item[$\bullet$] Given the set of input segments $S=\{s_i\}_{1\leq i\leq M}$ and the target time $T$, \textit{Random} will first produce a random integer $1\leq r\leq M$. Then it will randomly pick up $r$ segments from $S$ to get the result $A=\{a_i\}_{1\leq i\leq r}$, regardless of the target time $T$.
    \item[$\bullet$] Given $S=\{s_i\}_{1\leq i\leq M}$ and $T$, \textit{Random-Cut} randomly picks up segments from $S$ and add to $A$ until $c_1\cdot T\leq\tau(A)\leq c_2\cdot T$, ensuring satisfying the requirement of duration.
    \item[$\bullet$] Given $S=\{s_i\}_{1\le i\leq M}$ and $T$, the \textit{SAM} \cite{MVSM} will utilize an extra model to perform video classification or named entity recognition to obtain some labels for each segment and compute an importance score for each segment. It also utilizes an extra BERT \cite{bert} to perform next sentence prediction (NSP) to compute a coherence score for each pair of segments. Then the segments $\{s_i\}_{1\leq i\leq M}$ will be modeled as a graph with $|S|$ nodes and $|S|(|S|-1)$ edges, where the weight of nodes are the importance score and the weight of edges are the coherence score. DFS with pruning is then adopted to search on the graph to collect a set that maximizes the sum of importance scores and coherence scores.
\end{itemize}
\subsubsection{Parameters.}
The Swin-Transformer \cite{swin} we used is the Large version with an output size of 1536. The 5 shots will be extracted to generate a visual representation every second of the video. The segment-level representation is computed as the mean of. The output size of BERT \cite{bert} and Vggish \cite{vggish} are 768 and 128 respectively, and the sampling rate of Vggish is 5 every second to align with visual information. The dimension of $\tilde s_i$ is 2432. The linear embedding layer performs a  projection from 2432 to 768. We use a two-layer Bi-GRU with hidden size of 256 as encoder and one-layer GRU with hidden size of 512 as decoder.
\subsubsection{Training Details.}
We optimize the sum reward $R=0.5\cdot R_{imp}+0.5\cdot R_{coh}$, where the $R_{imp}$ and $R_{coh}$ are given in Eq. \ref{eq:def_R_imp} and Eq. \ref{eq:def_R_coh} with $\beta=0.5$. The $w_l$ in $R_{imp}$ for all $\ell$ are 0.25. The optimizer we used is Adam \cite{adam}. There are 8 videos in each batch, and the learning rate $\eta=2\times 10^{-4}$. The number of episodes for each video $K=8$, and the number of epochs is 10. The end segment of every ad video usually contains a wealth of ad-related information. Therefore, the end segment will be selected to be the first item of $A$ in our implementation. More details can be found in supplementary.
\subsection{Performance Comparison}
We evaluated our M-SAN on the test set of Ads-1k with the three baselines mentioned above. The results are provided by Tbl. \ref{tb:cmp}. It shows that our M-SAN is state of the art on segment assemblage task given target duration $T=10$ and $T=15$. There is a significant improvement from Random to Random-Cut. Therefore, simply sticking to the time limit can improve performance by leaps and bounds. From Random-Cut to SAM, the improvement at $T=10$ is also obvious, while the difference between them at $T=15$ is not. This is probably because a longer duration budget forces Random-Cut to select more segments. The segment pairs as result have a greater chance of being coherent pairs.

Although SAM performs better than Random-Cut, its ability to trading-off between importance and coherence is still weak. M-SAN addresses the problem by trained with importance-coherence reward and achieves a better performance.
\vspace{-1.5em}
\begin{table}
\begin{center}
\caption{Performance comparison results. Our M-SAN is state of the art on segment assemblage task.}
\label{tb:cmp}
\begin{tabular}{c | c  c  c | c  c  c}
\hline
&\multicolumn{3}{c|}{Imp-Coh@10}&\multicolumn{3}{c}{Imp-Coh@15}\\
\cline{2-7}
  & ~~~Imp~~~ & ~~~Coh~~~ & ~Overall~ & ~~~Imp~~~ & ~~~Coh~~~ & ~Overall~ \\ 
\hline
Random & 10.68 & 12.57 & 7.92 & 17.04 & 20.49 & 12.71\\  
Random-Cut & 55.35 & 73.75 & 41.77 & 60.39 & 76.56 & 47.25\\
SAM \cite{MVSM} & 72.54 & 86.55 & 65.97 & 63.97 & 78.58 & 58.09\\
\textbf{M-SAN(ours)} & \textbf{80.29}	&\textbf{92.16}&	\textbf{74.19}&	\textbf{77.00}&	\textbf{90.83}&	\textbf{70.28}\\
\hline
\end{tabular}
\end{center}

\end{table}
\vspace{-3.5em}
\subsection{Ablation Studies}
\subsubsection{Ablation of Modalities.}
We explore the effect of the representation from different modalities. Tbl. \ref{tb:modals} shows that incorporating the text or audio representation can both improve the overall score while incorporating the former have much effect. After leveraging all representations from three modalities, the overall scores increase significantly, which has 8.7 and 3.75 gains compared with utilizing visual information only on Imp-Coh@10 and Imp-Coh@15, respectively. Even though adding text representation to video-audio dual modalities hurts the Imp@15, other scores increase obviously, which demonstrates the significance of multi-modal representation.
\begin{table}
\begin{center}
\caption{Ablation study on modalities}
\label{tb:modals}
\begin{tabular}{ c  c  c | c  c  c | c  c  c }
\hline
\multicolumn{3}{c|}{Modalities}&\multicolumn{3}{c|}{Imp-Coh@10}&\multicolumn{3}{c}{Imp-Coh@15}\\
\hline
 ~V~&~T&A&Imp~&~Coh~&~Overall~&~Imp~&~Coh~&~Overall~\\ 
\hline
\checkmark & ~ & ~ & ~78.43~ & ~89.59~ & ~65.49~ & ~74.48~ & ~87.41~ & ~66.53~ \\
\checkmark & \checkmark & ~ & ~78.57~ & ~89.73~ & ~71.04~ & ~76.22~ & ~89.85~ & ~68.56~ \\
\checkmark & ~ & \checkmark & ~79.49~ & ~91.49~ & 72.72 & \textbf{77.08} & 90.36 & 69.91 \\
~\checkmark~ & ~\checkmark~ & ~\checkmark~ & ~\textbf{80.29}~&~\textbf{92.16}~&~\textbf{74.19}~& ~77.00~&~\textbf{90.83}~&~\textbf{70.28}~\\
\hline
\end{tabular}
\end{center}

\end{table}

\subsubsection{Ablation of Reward and Glimpse.}
To verify the effectiveness of reward and glimpse (two-stage attention calculation), we design the following ablation experiments with target duration $T=10$ and $T=15$. 
The results in the Tbl. \ref{rw_g} show that M-SAN gains a higher score on all metrics than the one without glimpse. Therefore, dynamically integrating information over the whole video by Glimpse can improve the performance.
Similarly, we perform ablation on the rewards: importance-coherence reward (M-SAN), importance reward only, and coherence reward only. Results in Tbl. \ref{rw_g} show that M-SAN trained with importance-reward only gained relatively low scores comparing the other two kinds of rewards. The one trained with coherence-reward only gained a higher score than the one with importance-reward. M-SAN trained with importance-coherence reward prominently performs better than the other two.
\vspace{-1.0em}
\begin{table}
\begin{center}
\caption{Ablation study on glimpse and rewards.}
\label{rw_g}
\begin{tabular}{c| c  c  c |  c  c  c }
\hline
&\multicolumn{3}{c|}{Imp-Coh@10}&\multicolumn{3}{c}{Imp-Coh@15}\\
\cline{2-7}
& Imp&Coh & Overall & Imp & Coh & Overall\\ 
\hline
\textbf{M-SAN}& \textbf{80.29} & \textbf{92.16} & \textbf{74.19} & \textbf{77.00} & \textbf{90.83} & \textbf{70.28} \\
w/o glimpse&79.80&91.50&72.97&75.98&90.58 & 70.12\\
coh-rwd only& 79.37~&~91.02 & 72.28 & 76.90&90.29 & 69.75\\
imp-rwd only& 67.82&71.62 & 49.87 & 67.42&79.89 & 54.10\\  
\hline
\end{tabular}
\end{center}
\end{table}
\vspace{-3.5em}
\subsubsection{Analysis of Reward Ratio.}
\begin{figure}
\centering
\includegraphics[width=120mm]{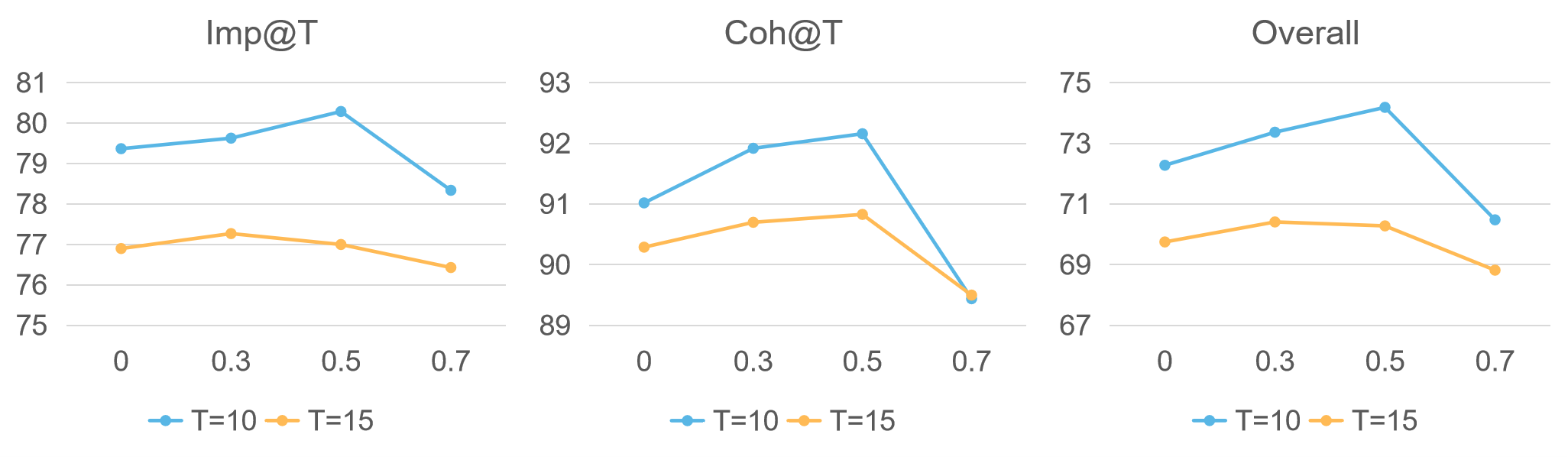}
\vspace{-0.5em}
\caption{Comparison results with $\beta=0.0/0.3/0.5/0.7$.}
\label{fig:chart}
\end{figure}

To further figure out which reward ratio brings the optimal results, we experiment on $T=10$ and $T=15$, assigning 0.0/0.3/0.5/0.7 to $\beta$. Importance score, coherence score, and overall score results are shown in the line chart Fig. \ref{fig:chart}. The results at the $T=10$ and $T=15$ are presented by lines painted in blue and orange respectively. Given target duration $T=10s$, all three scores reach a peak at $\beta$=0.5. Given target duration $T=15s$, the importance score and overall score reach a peak at $\beta$=0.3, which is slightly higher than the score at $\beta$=0.5. Coherence score reaches the highest point at $\beta$=0.5. On the whole, the performance is relatively good at the condition of $\beta$=0.5.

\subsection{Qualitative Analysis}
\begin{figure}
\centering
\includegraphics[width=115mm]{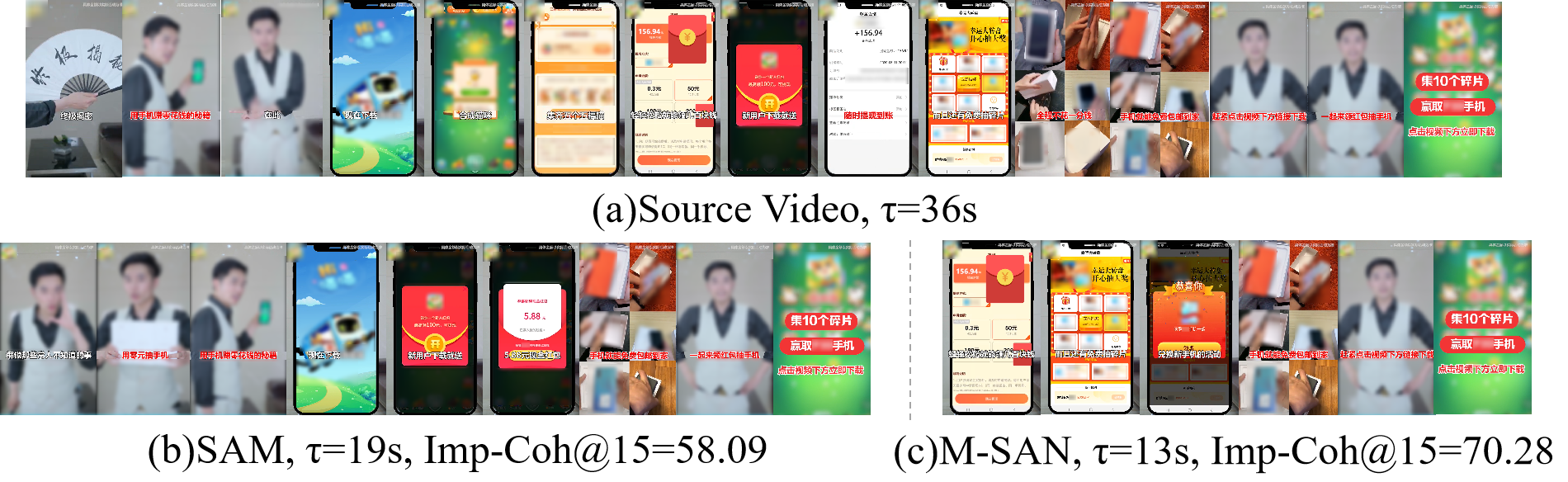}
\caption{Visualization. Source video and videos assembled by SAM and M-SAN given target duration $T=15s$. $\tau$ is the actual duration of result.}
\label{fig:15}
\end{figure}

One result is shown as Fig. \ref{fig:15}. The source video is about a collectible game app with real mobile phones as the completion rewards. Its original duration is 36s, and the target duration is 15s. SAM \cite{MVSM} generates videos with too many foreshadowing parts, which exceeds the duration limitation. M-SAN produces a 15s result and tends to select latter segments in source videos, which is reasonable because key points usually appear in the latter part of the ad with the front part doing foreshadowing. The result of M-SAN first demonstrates using the app to get a new phone and then shows a scene where a new phone is packed, which emphasizes the rewards of completing the game. This verifies the M-SAN focuses on more informative segments and does better than SAM in duration control.
\vspace{-2.0em}
\section{Conclusion}

The two main stages of ad video editing are video segmentation and segment assemblage. Existing methods perform poorly at the segment assemblage stage. To improve the performance of segment assemblage, we proposed M-SAN to perform segment assemblage end-to-end. We also proposed importance-coherence reward based on the characteristics of ad and train M-SAN with policy gradient. We collected an ad video dataset with 1000+ ad videos and proposed Imp-Coh@Time metrics. Experimental results show the effectiveness of M-SAN and verify that multi-modal representation and importance-coherence reward bring a significant performance boost.

\section*{Acknowledgment}
This work is supported by the National Natural Science Foundation of China under Grant No. 61972188 and 62122035.
%
%
\bibliographystyle{splncs04}
%


\clearpage
\appendix
\section{Annotation Methods}
\label{annotation}
\subsection{Narrative Technique Annotation}
Narrative label is defined as the description of the techniques used in the development of videos. In the advertisement area, labels can be divided into 5 basic categories: background foreshadowing, sore points, product display, brand reinforcement, and behaviour guidance. Under these five categories, there are more specific labels up to 81. The labels are finally divided into 4 groups according to their importance and listed in Tbl. \ref{labellist}.

There are a few rules on narrative technique annotation:
\begin{itemize}
    \item[$\bullet$] One segment may be assigned a few labels.
    \item[$\bullet$] The overall order of labels of segments from the same source video is background foreshadowing, sore points, product display/brand reinforcement, and behaviour guidance roughly. The order may differ but the last segment should not have the label of background foreshadowing.
    \item[$\bullet$] Those labels have industrial characteristics.
\end{itemize}

\begin{table}
\begin{center}
\caption{Narrative technique labels. They are divided into 4 groups. The importance scores of a single segment are computed as the weighted sum of group-level weights of all labels assigned, where the weights are the fourth powers of 0.25, 0.5, 0.75, and 1.00 for group-level 1, 2, 3, and 4 respectively in our implementation.}
\label{labellist}
\begin{tabular}{| c | c | c |}
\hline
~Group~ & \multicolumn{2}{c|}{Labels}\\\hline
& ~~~~~~~~~plain and ungarnished~~~~~~~~~ & contrasts \\\cline{2-3}
 1& question and suspense & ~~~~~emotional resonance~~~~~~ \\\cline{2-3}
 & background forshadowing & -\\\hline

& personal statement & celebrity introduction \\\cline{2-3}
 & releasing notices & dialogue opening \\\cline{2-3}
 & narrator statement & merchandise opening \\\cline{2-3}
 & apology & longitudinal comparison of characters \\\cline{2-3}
& oops & behaviour comparison \\\cline{2-3}
& product comparison & rhetorical question \\\cline{2-3}
2& conflict question & novelty elements \\\cline{2-3}
& plot anticipation & music anticipation \\\cline{2-3}
& factual description & horizontal comparison of characters \\\cline{2-3}
& target population & realistic \\\cline{2-3}
& yearning description & anxiety creation \\\cline{2-3}
& guarantee & repeatedly emphasis \\\cline{2-3}
& good fronting & sore points \\\hline

& question rhetorical class & aiming at target population \\\cline{2-3}
& aiming at age grades & aiming at behaviour characteristics \\\cline{2-3}
& time need & demand description technique \\\cline{2-3}
& status need & financial incoming \\\cline{2-3}
3& insufficient fund & aiming at application scenarios \\\cline{2-3}
& living needs & entertainment needs \\\cline{2-3}
& health need & relationship needs \\\cline{2-3}
& job needs & learning needs \\\cline{2-3}
& maintenance needs & requirements type \\\hline


& product function display & ~~~~~~~~~product quality display~~~~~~~~~ \\\cline{2-3}
& entire product display & product details display \\\cline{2-3}
& product usage display & model display \\\cline{2-3}
& product advantage display & product display(other) \\\cline{2-3}
& environment display & business service display \\\cline{2-3}
& business effect display & business promotion display \\\cline{2-3}
& business process display & business advantages display \\\cline{2-3}
4& business display(other) & reading image \\\cline{2-3}
& short video apps & live streaming picture \\\cline{2-3}
& withdraw picture & operational guideline \\\cline{2-3}
& merchandise display & application display(other) \\\cline{2-3}
& combat playing methods & interesting display method \\\cline{2-3}
& social playing method & painting style display \\\cline{2-3}
& character display & selling point display \\\cline{2-3}
& equipment display & playing method display \\\cline{2-3}
& armament display & game withdraw \\\hline
\end{tabular}
\end{center}
\end{table}

\subsection{Coherence Annotation}
In order to evaluate the coherence of the model output, we annotate coherence labels for the testing set. Specifically, we combine every two adjacent segments in the result with $N$ segments to yield $N\cdot(N-1)$ pairs. Then the annotators will assign labels to these pairs with the following method:
\begin{itemize}
    \item[$\bullet$] If the annotators think that the segments in pairs are connected coherently, then the label is assigned as \textit{coherent}.
    \item[$\bullet$] If the annotators think that the segments in the pair are connected incoherently, then the label is assigned as \textit{incoherent}.
    \item[$\bullet$] Otherwise, the label is assigned as \textit{uncertain}.
\end{itemize}

\section{Visualization}
\label{quali-ana}
\begin{figure}[h]
\centering
\includegraphics[width=110mm]{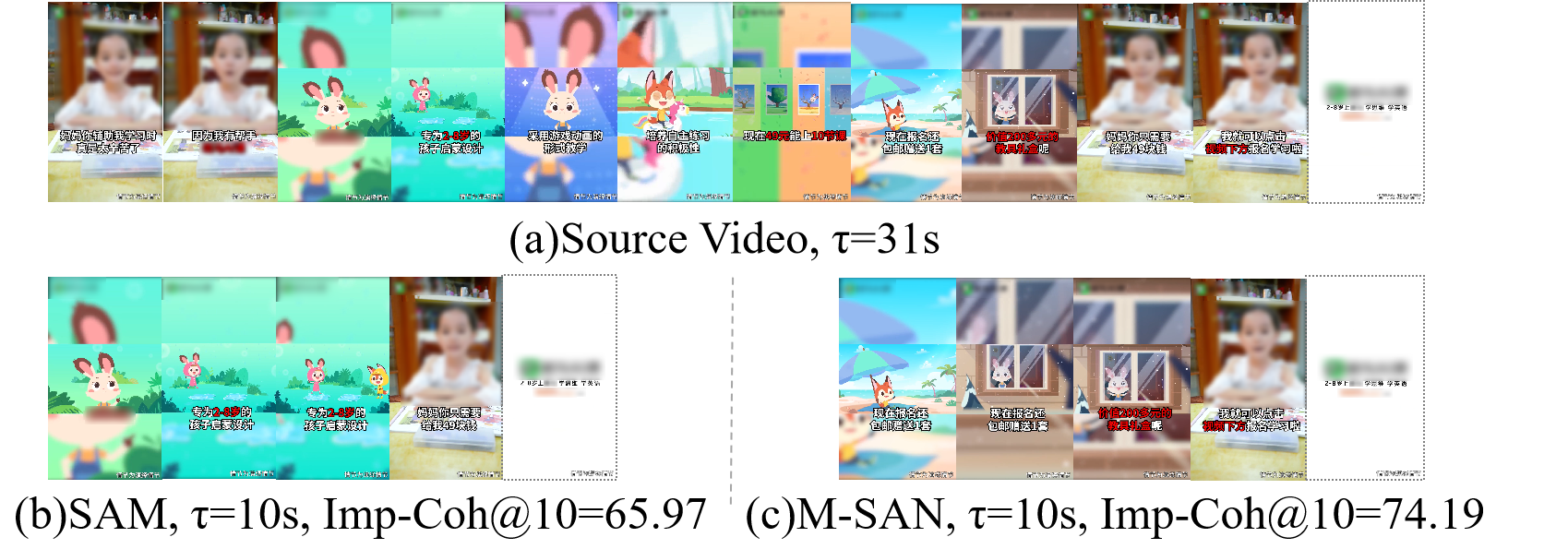}
\caption{Source video and videos assembled by SAM and M-SAN with a given target duration $T=10s$. $\tau$ is the actual duration of the result.}
\label{fig:10}
\end{figure}

Fig. \ref{fig:10} presents a source video and videos assembled by SAM and M-SAN in a target duration of 10s. The content details of these videos are listed below.\\
\textbf{Source Video}(a): A little girl told her mum there’s no need to do homework tutoring for her. Because there’s a tutorial course designed for children aged 2-8. This course adopts instructional design using game animation, aiming to cultivate the enthusiasm for exercise initiatively. Now it is available for ten lessons for 49 yuan. If you apply now, a gift of teaching aid gift box worth more than 200 yuan will be sent to you. Then the little girl told her mom to give her 49 yuan to pay for the course. Then she will be available to click the link below and study! The final segment is the company logo presentation of this course.\\
\textbf{SAM}(b): There’s a tutorial course designed for children aged 2-8. A little girl told her mom to give her 49 yuan to pay for the course.  The final segment is the company logo presentation of this course.\\
\textbf{M-SAN}(c): There’s a tutorial course for which if you apply now, a gift of teaching aid gift box worth more than 200 yuan will be sent to you. A little girl said that she will be available to click the link below and study! The final segment is the company logo presentation of this course.

\begin{figure}
\centering
\includegraphics[width=115mm]{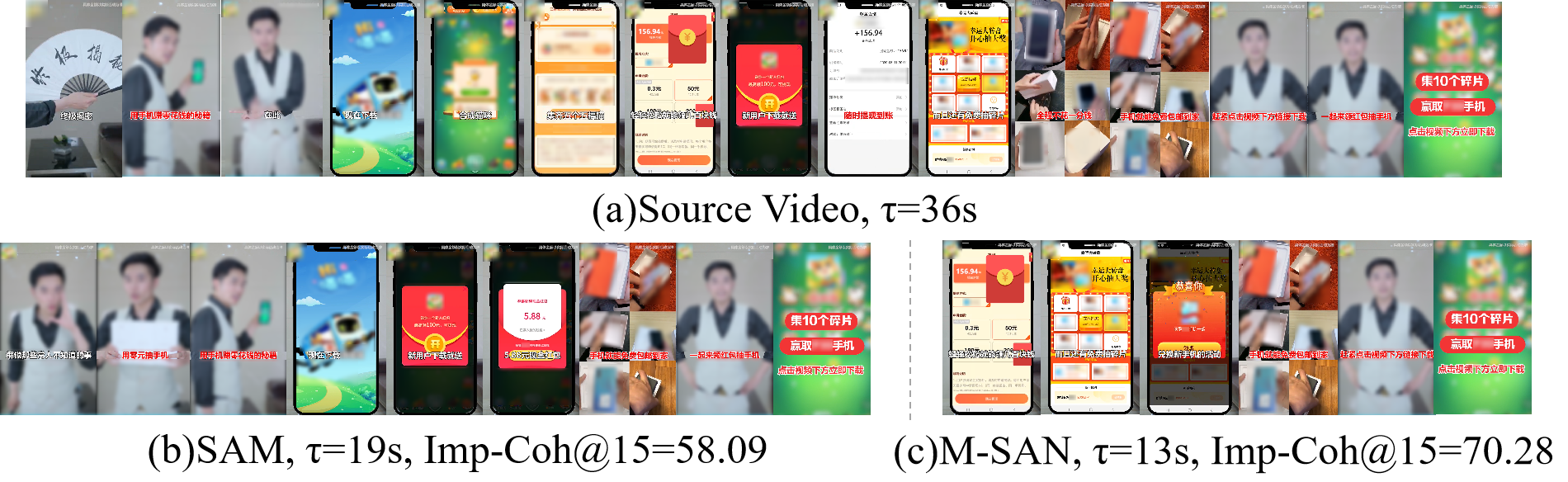}
\caption{Source video and videos assembled by SAM and M-SAN with a given target duration $T=15s$. $\tau$ is the actual duration of the result.}
\label{fig:15-2}
\end{figure}

Fig. \ref{fig:15-2} presents a source video and videos assembled by SAM and M-SAN in a target duration of 15s. The content details of these videos are listed below.\\
\textbf{Source Video}(a): The ultimate secret to earning pocket money from your phone is here! Now download this app, and synthesize cats to collect five blessing cats. Then you will get hundreds of money easily. Money will be given to new users once the app got downloaded. Withdraw cash at any time. And there’s a free lottery. It doesn't cost a penny, and you can get a new phone freely without paying any postage. Click on the link below and download this app. Let’s grab red envelopes and get a new phone! The final segment is the company logo presentation of this app.\\
\textbf{SAM}(b): The ultimate secret to earning pocket money from your phone is here! Now download the app. A red envelope of 5.88 yuan will be given to new users once the app got downloaded. It doesn't cost a penny, and you can get a new phone freely without paying any postage. Let’s grab red envelopes and get a new phone! The final segment is the company logo presentation of this app.\\
\textbf{M-SAN}(c): You can get hundreds of money easily. And there’s a free lottery for exchanging for a new mobile phone. It doesn't cost a penny, and you can get a new phone freely without paying any postage. Click on the link below and download this app. The final segment is the company logo presentation of this app.

It can be observed that SAM generates videos with too many foreshadowing parts, which exceeds the duration limitation. M-SAN tends to select latter segments in source videos, which is reasonable because key points usually appear in the latter part of the ad with the front part doing foreshadowing. This verifies the M-SAN focuses on more informative segments and does better than SAM in duration control.

\section{User Study}
Since we have the intention of deploying the model to produce ads in our online services, we had already done a user study. We used the test set as the input of both SAM  \cite{MVSM} and M-SAN and invited 6 colleagues from the advertising business department to evaluate the usability (usable or not usable) of the 15s results by subjectively judging whether the content was coherence and retained important commercial information and whether the output met the requirement of duration. The study shows the usability rate (\#usable results/\#all results) of M-SAN's output is 0.859, and the usability rate of SAM's output is 0.616. And the usability obtained from users is consistent with the evaluation of our metrics.
\begin{table}[h]
\begin{center}
\small
\vspace{-2.0em}
\renewcommand\arraystretch{1.0}
\setlength{\tabcolsep}{1mm}{
\begin{tabular}{l|c|c}
    \hline
    Methods & Usability & Imp-Coh@15\\
    \hline
    SAM  \cite{MVSM} & 0.616 & 58.09\\
    M-SAN (ours) & 0.859 & 70.28\\
    \hline
\end{tabular}}
\end{center}

\caption{User study.}

\label{table:abl}
\end{table}

\section{More Training and Testing Details}
The large models Swin-Transformer and BERT are frozen. And we fine-tuned GPT-2 that computes PPL on 8 A100 GPUs for 4 days. We trained M-SAN on 4 Tesla T4 GPUs for one day. The testing on 99 videos only needed 2 minutes on single Tesla T4 GPU.

\end{document}